%% file: main.tex
\pdfoutput=1

\documentclass[11pt]{article}

\usepackage{acl}

\usepackage{times}
\usepackage{latexsym}

\usepackage[T1]{fontenc}

\usepackage[utf8]{inputenc}

\usepackage{microtype}

\usepackage{inconsolata}
\usepackage{multirow}
\usepackage{enumitem}
\usepackage{color, colortbl}
\usepackage{xcolor}
\usepackage{multicol}
\usepackage{listings}
\definecolor{myblue}{RGB}{0,0,255}
\definecolor{myorange}{RGB}{255,165,0}

\title{Limits of Theory of Mind Modelling in\\Dialogue-Based Collaborative Plan Acquisition}

\newcommand\blfootnote[1]{%
  \begingroup
  \renewcommand\thefootnote{}\footnote{#1}%
  \addtocounter{footnote}{-1}%
  \endgroup
}

\author{Matteo Bortoletto \quad Constantin Ruhdorfer$^{\dag}$ \quad Adnen Abdessaied$^{\dag}$ \\ \textbf{Lei Shi} \quad \textbf{Andreas Bulling} \\
  University of Stuttgart, Germany \\
  \texttt{matteo.bortoletto@vis.uni-stuttgart.de}}

\usepackage{amssymb}
\usepackage{amsmath}
\usepackage{booktabs}
\newtheorem{definition}{Definition}
\usepackage{graphicx}

\newcommand*{\img}[1]{%
    \raisebox{-.15\baselineskip}{%
        \includegraphics[
        height=\baselineskip,
        width=\baselineskip,
        keepaspectratio,
        ]{#1}%
    }%
}

\begin{document}

\maketitle

\begin{abstract}
    \input{latex/00_abstract}
\end{abstract}

\input{latex/01_intro}

\input{latex/02_related_work}

\input{latex/03_method}

\input{latex/04_experiments}

\input{latex/05_conclusion}
\input{latex/06_limitations}

\input{latex/07_ethics}

\bibliography{anthology,custom}

\clearpage
\appendix
\input{latex/08_appendix}

\end{document}

%% file: latex/00_abstract.tex
Recent work on dialogue-based collaborative plan acquisition (CPA) has suggested that Theory of Mind (ToM) modelling can improve missing knowledge prediction in settings with asymmetric skill sets and knowledge.
Although ToM was claimed to be important for effective collaboration, its real impact on this novel task remains under-explored.
By representing plans as graphs and exploiting task-specific constraints we show that, as performance on CPA nearly doubles when predicting one's own missing knowledge, the improvements due to ToM modelling diminish. 
This phenomenon persists even when evaluating existing baseline methods.
To better understand the relevance of ToM for CPA, we report a principled performance comparison of models with and without ToM features.
Results across different models and ablations consistently suggest that features learnt for ToM tasks are more likely to reflect latent patterns in the data with no perceivable link to ToM.
This finding calls for a deeper understanding of the role of ToM in CPA and beyond, as well as new methods for modelling and evaluating mental states in computational collaborative agents. 
\blfootnote{$^{\dag}$Equal second-author contribution.} 
\blfootnote{
\img{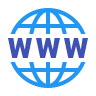} 
The project web page is accessible \href{https://perceptualui.org/publications/bortoletto24_acl/}{here}.
}

%% file: latex/01_intro.tex
\section{Introduction}
\label{sec:intro}

Dialogue-based human-AI collaboration is an interaction in which humans and artificial intelligent (AI) agents converse to achieve a shared goal or task~\citep{streeck2011embodied}.
When humans collaborate with each other, they rely on two main abilities: Verbal communication and Theory of Mind (ToM), i.e.\ the ability to infer one's own and others' mental states~\citep{premack1978does}. 
To succeed in collaborating with humans, it is therefore imperative for AI agents to possess similar capabilities \citep{williams2022supporting}.

\begin{figure}[t]
    \centering
    \includegraphics[width=\columnwidth]{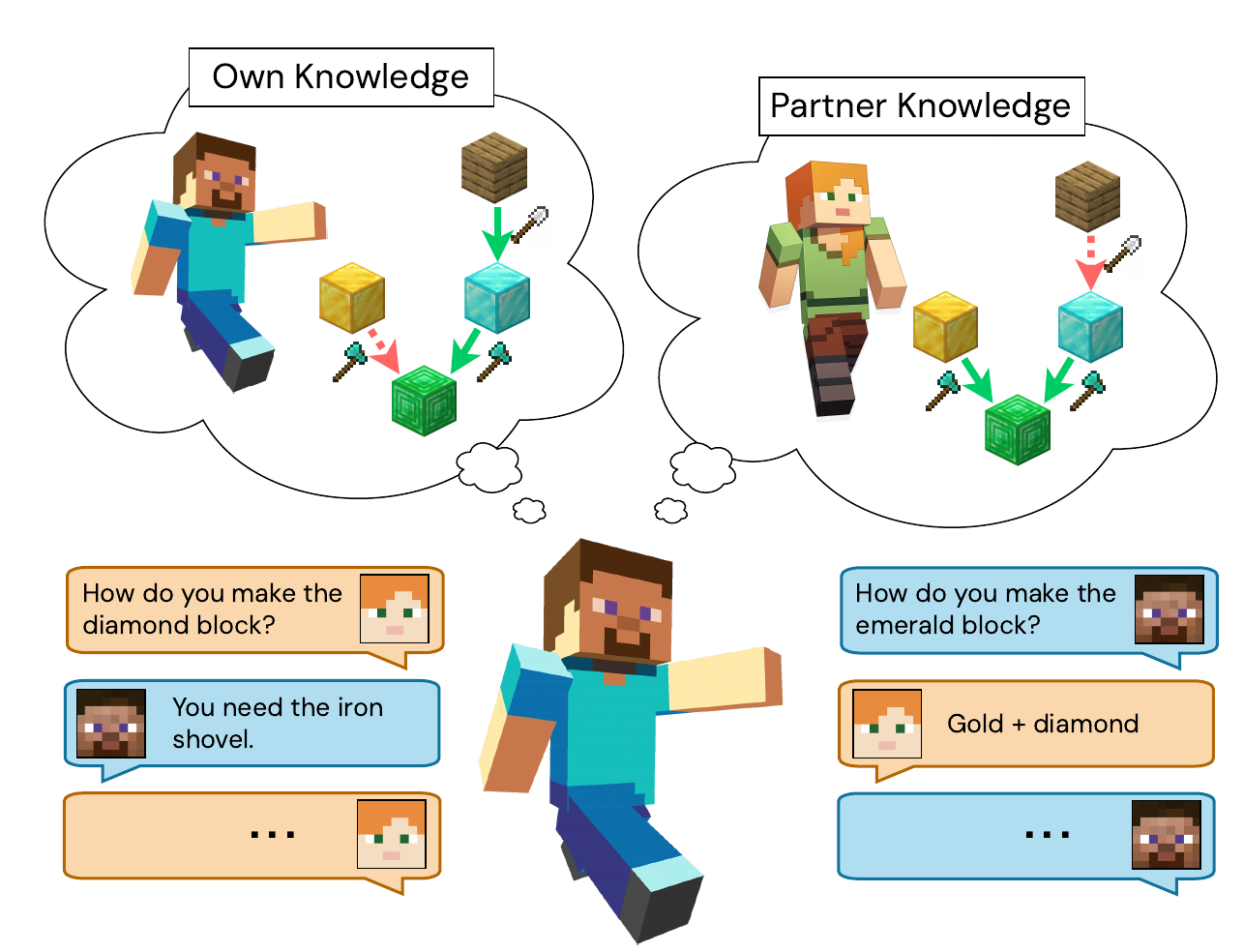}
    \caption{Collaborative plan acquisition in MindCraft involves inferring one's own and the partner's missing knowledge (\img{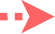}) through situated dialogue, starting from individual partial plans (\img{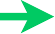}) to achieve a shared goal.}
    \label{fig:cpa}
\end{figure}

Recent work on this topic has introduced collaborative plan acquisition (CPA) as a promising task for evaluating collaborative abilities in agents \citep{bara2023towards}.
Starting from asymmetric knowledge and skill-sets of two collaborating agents, the goal of CPA is to infer one's own missing knowledge (OMK) and the partner's missing knowledge (PMK) to achieve a shared goal by engaging in a multi-round situated dialogue (see \autoref{fig:cpa}).
To study CPA, the authors used \textit{MindCraft} -- a multi-modal collaborative dialogue-based benchmark grounded in the sandbox game Minecraft \citep{bara-etal-2021-mindcraft}.
They also proposed a sequence-to-sequence CPA model
that used visual observations, plan, and dialogue history as input.
Their empirical results showed a large difference in performance between predicting OMK and PMK.
Furthermore, they found that while including a subset of ToM features improved performance, using all ToM features resulted in nearly the same performance as using none.

In this work, we systematically analyse the limits of ToM modelling in dialogue-based collaborative plan acquisition.
We first propose to represent plans as directed graphs, each represented by a node feature matrix, a connectivity matrix, and an edge feature matrix.
This starkly contrasts previous works \cite{bara-etal-2021-mindcraft, bara2023towards} that represented plans as lists consisting of materials and tools processed by GRU~\cite{cho-etal-2014-learning}.
Our novel structured representation allows us
to elegantly frame CPA as a link prediction task~\citep{liben2003link} and apply negative sampling 
constrained on \textit{MindCraft} plans' structure for efficient training.
Our proposed representation not only doubles the performance of predicting OMK compared to \cite{bara2023towards} but also bridges the gap to predicting PMK. 

However, our evaluations show no significant performance difference when using ToM features -- neither for our method nor for the baselines
from \citet{bara2023towards}.
We thus conduct extensive analyses using diagnostic probing, correlation analysis, and ground-truth ToM labels as input.
Results across different models and ablations consistently suggest that learned ToM features are less associated with mental states and more aligned with revealing latent patterns within the data.

In summary, the contributions of our work are two-fold:
(1) We propose a novel graph-based representation of plans for CPA and show that applying graph learning methods simultaneously doubles the performance of predicting OMK 
and closes the gap to predicting PMK;
(2) We report principled analyses
across different models and ablations that suggest that learnt ToM features reflect latent patterns in the data with no perceivable link to ToM.

%% file: latex/02_related_work.tex
\section{Related Work}
\label{sec:rel_work}

\subsection{Dialogue-based Human-AI Collaboration}

Collaborative dialogue systems are designed to work with humans towards achieving a shared goal 
\citep{rich2001Collagen,bohus2009the,allen2002A, streeck2011embodied}.
Early works were based on scripts~\citep{Traum2017Computational}, employed planning~\citep{Papaioannou2018HumanRobotIR}, or modelled the dialogue as a collection of information states \citep{Larsson2000Information}.
More recent work focused on neural sequence-to-sequence models to learn from dialogue corpora~\citep{wen2015semantically,dong2022A}.
Neural approaches have also been explored for collaborative dialogues taking place when participants are working on a shared artefact within a co-observed environment~\cite{narayan-chen-etal-2019-collaborative, kim-etal-2019-codraw, jayannavar-etal-2020-learning, bara-etal-2021-mindcraft, bara2023towards}.
Another line of work explored the role of Theory of Mind~\citep{premack1978does} in dialogue-based collaboration, focusing on simulated textual environments~\citep{qiu-etal-2022-towards, zhou-etal-2023-cast}, or on human gameplay~\cite{bara2023towards}.

Our work focuses on the \textit{MindCraft} environment~\cite{bara-etal-2021-mindcraft, bara2023towards} in which two agents with asymmetric skill sets and knowledge converse to achieve a shared goal in a Minecraft world. 

\begin{figure*}[t]
    \centering
    \includegraphics[width=\linewidth]{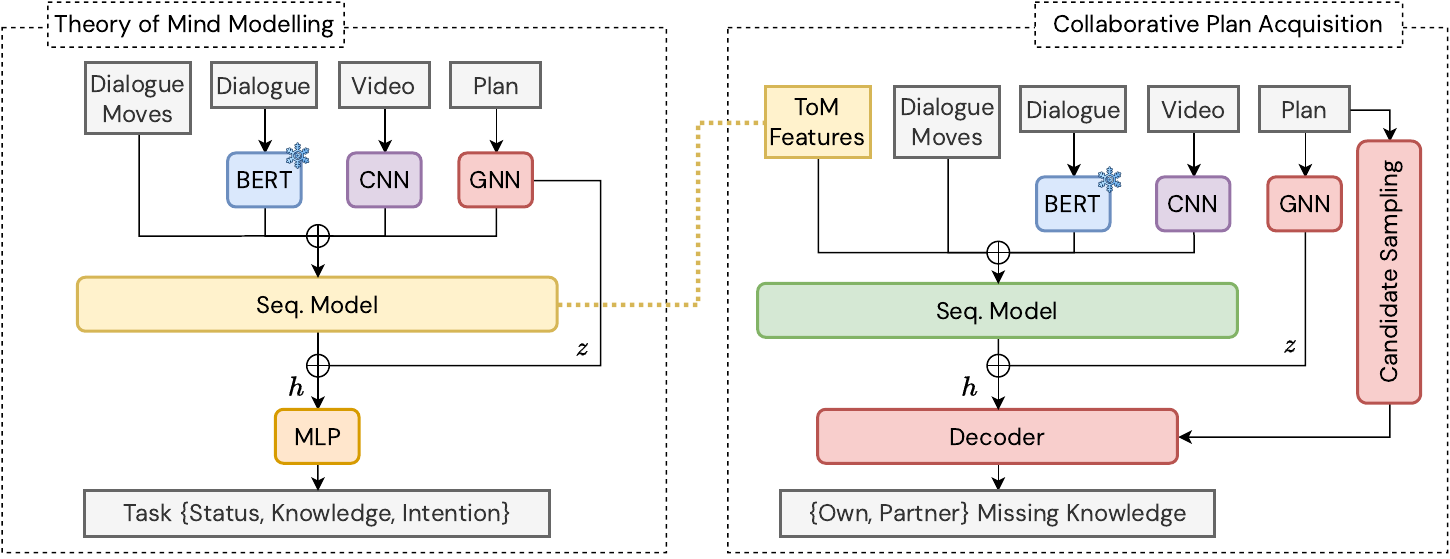}
    \caption{Models' architecture for Theory of Mind (ToM) modelling and collaborative plan acquisition (CPA). Following \citep{bara-etal-2021-mindcraft, bara2023towards}, we train one model for each ToM task (\textit{Status}, \textit{Knowledge}, \textit{Intention}) and CPA task (Own Missing Knowledge, Partner's Missing Knowledge) and freeze the BERT weights during training (indicated by \img{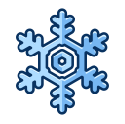}) for a fair comparison with the baseline.}
    \label{fig:model}
\end{figure*}

\subsection{Computational Theory of Mind}
With recent advances in AI, an increasing number of works studied means to equip models with Theory of Mind capabilities based on deep learning approaches~\cite{rabinowitz2018machine, bara-etal-2021-mindcraft, gandhi2021baby, zhou-etal-2023-cast, liu2023computational,bortoletto24_aaai}, partially observable Markov decision processes~\cite{doshi2010modeling, han2018learning} or via Bayesian approaches~\cite{baker2009action, lee2019bayesian, buehler2020theory, fan2021learning}.
Within these, one line of work focused on inferring beliefs, actions, or instructions solely \textit{as an observer} of agent behaviour~\citep{rabinowitz2018machine,grant2017can,duan2022boss}.
An emerging second line of work explored ToM \textit{from the perspective} of interacting agents \cite{Wang2021Towards,qiu-etal-2022-towards,bara2023towards}, 
highlighting the significance of ToM in collaborative tasks. 
\citet{bara2023towards} claimed that integrating ToM features improves collaborative plan acquisition (CPA). 
However, the limits and failure cases of ToM-enabled agents are poorly understood, particularly whether they really model mental states or exploit dataset biases. 
This work aims to address these concerns and assess ToM modelling on CPA.

%% file: latex/03_method.tex
\section{Problem Formulation}
\label{sec:prob_form}
\begin{figure*}[t]
    \centering
    \includegraphics[width=\linewidth]{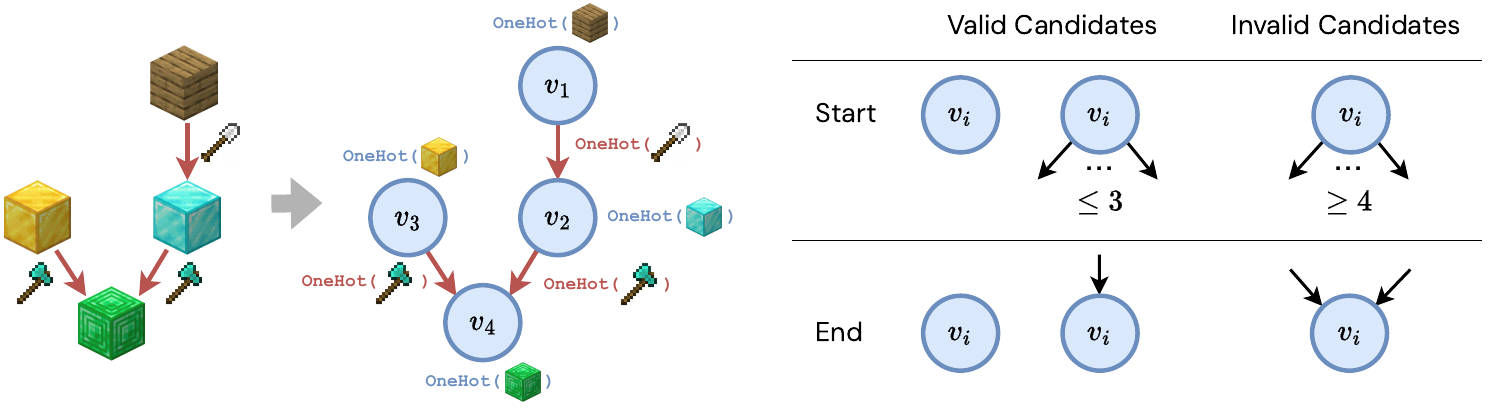}
    \caption{\textbf{Left:} Example of a plan graph. 
    Nodes represent materials, edges connect each material to the requisite components for its synthesis, and edge features denote the tool needed to interact with a material.
    \textbf{Right:} Our candidate sampling strategy for predicting OMK: Valid candidate start-nodes must have an out-degree smaller than four whereas candidate end-nodes must have an in-degree less than or equal to one.
    }
    \label{fig:graph_and_candidate}
\end{figure*}

\textit{MindCraft} \cite{bara-etal-2021-mindcraft} was introduced as a multi-modal benchmark for studying ToM modelling within collaborative tasks.
It involves two players collaborating through dialogue in a 3D block world to craft a target material by manipulating blocks using specific tools (see \autoref{fig:graph_and_candidate}, left).
Players initially receive a partial plan as an incomplete directed AND-graph and a tool allowing each to interact with a set of specific blocks.
Endowed with complementary knowledge and skill sets, players must communicate via the in-game chat to craft the target material and reason about each other's mental states. 
The ToM tasks introduced by~\citet{bara-etal-2021-mindcraft} are specifically designed to capture mental state information pertinent to collaboration. 
During gameplay, players are presented with pop-up questions every 75 seconds, each paired by type:
\begin{enumerate}[leftmargin=*, itemsep=0pt]
    \item \textit{Task Status}: Predict if one of the two players has created a specific material. 
    For instance, Player 1 is asked: ``\texttt{Has your partner created GOLD\_BLOCK so far?}'' and Player 2 is asked: ``\texttt{Have you crafted GOLD\_BLOCK yet?}''
    Possible answers are \texttt{YES}, \texttt{NO}, or \texttt{MAYBE}.
    \item \textit{Player Knowledge}: Predict whether players know how to craft a material or if they believe their partner knows. 
    For example, Player 1 is asked: ``\texttt{Do you think the other player knows how to make BLUE\_WOOL?}'' and Player 2 is asked: ``\texttt{Do you know how to make BLUE\_WOOL?}''
    Possible answers are \texttt{YES}, \texttt{NO}, or \texttt{MAYBE}. 
    \item \textit{Player Intention}: Predict which material a player is making at the current time step. 
    For example, Player 1 is asked: ``\texttt{What do you think the other player is making right now?}'' and Player 2 is asked: ``\texttt{What are you making right now?}''. 
    Possible answers are the different types of blocks in the game or \texttt{NOT\_SURE}. 
\end{enumerate}

In this work, we focus on a recent extension of \textit{MindCraft} by \citet{bara2023towards} in which they proposed collaborative plan acquisition (CPA) and explored the role of ToM modelling in predicting players' missing knowledge while executing the crafting tasks. 
CPA is formulated as follows:
\begin{definition}\label{def:cpa}
    Consider a joint plan as a directed AND-graph $\mathcal{P} = (V,E)$, where the nodes $V$ denote (sub-)goal materials, and edges $E$  denote temporal constraints between the sub-goals. In a collaborative plan acquisition problem, two agents $i$ and $j$ start with partial plans $\mathcal{P}_i = (V, E_i)$, $E_i \subseteq E$, and $\mathcal{P}_j = (V, E_j)$, $E_j \subseteq E$. Given a sequence of visual observations $O^t_i$ and a joint dialogue history $D^t$ at time $t$, agent $i$ has to infer their own missing knowledge $\Bar{E}_i = E \setminus E_i$ and the partner $j$’s missing knowledge $\Bar{E}_j = E \setminus E_j$.
\end{definition}
ToM and CPA tasks are closely related but fundamentally different: ToM tasks focus on exploring players' \textit{beliefs} about the game state and their partner's mental states, while CPA tasks involve predicting \textit{missing information} from players' partial plans.
Additional details are in \S\ref{app:mc1} and \S\ref{app:mc2}.

\section{Method}
\label{sec:method}

\subsection{Baseline}
\label{sec:computational_model}

As a baseline we used the model proposed by \citet{bara2023towards} that embeds dialogue utterances using a frozen pre-trained BERT model \citep{devlin-etal-2019-bert} and represents each utterance by the features obtained from processing the corresponding \texttt{[CLS]} token using linear layers with $\tanh$ activation. 
A CNN and GRU were used to process the video frames and partial plans, respectively. 
During training, a model was first trained for each of the three ToM tasks of \S\ref{sec:prob_form} as classification tasks, requiring them to predict players' answers.
Afterwards, two separate models were trained on OMK and PMK by feeding the concatenation of the input modalities and the learnt ToM features into an LSTM~\citep{hochreiter1997long} followed by an MLP that outputs softmax scores for each possible missing link.

\subsection{GNN-based Missing Knowledge Prediction}
\paragraph{Representing Plans as Graphs.} 
We propose a modification to the method by \cite{bara2023towards} that includes representing plans as graph objects and using a GNN-based encoder-decoder together with candidate sampling to predict missing edges (see \autoref{fig:model}).
More specifically, as shown in \autoref{fig:graph_and_candidate}, nodes represent materials,
edges connect each material to the prerequisite components for its synthesis, and edge features denote the specific tool needed to interact with the material.
This structured representation allows us to more naturally represent the dependencies between materials and information about the tools involved in the crafting process.
Most importantly, it allows us to elegantly frame CPA as a link prediction task.

\paragraph{Theory of Mind Modelling.} 
We modified the ToM modelling architecture of \citet{bara-etal-2021-mindcraft} with two key changes: Replacing the GRU plan encoder with GATv2 layers~\citep{brody2022how} and average graph pooling, and substituting the LSTM with a single-block Transformer~\citep{vaswani2017attention}. The rest of the architecture remains unchanged for a fair comparison.

\paragraph{Missing Knowledge Prediction.} 
By representing plans as graphs we can perform missing knowledge prediction by applying negative sampling -- a common technique used for graph completion~\citep{yang2014embedding, schlichtkrull2018modeling}.
A GNN encoder $g$ first maps each node $v_i \in V$ to a real-valued vector $z_i = g(v_i) \in \mathbb{R}^d$. 
We then add $\Omega$ negative edges to the original graph. 
In conventional link prediction, negative edges are sampled randomly starting from the complete graph, and the goal is to classify edges as true or fake.
In contrast, our approach begins with an incomplete graph and the task is to predict missing edges. 
Randomly selecting negative edges poses the risk of missing the edges we aim to predict.
Our improved approach uses the plan structural constraints of \textit{MindCraft} to narrow down the pool of all possible $V^2 \backslash E_{\{i,j\}}$ edges to a set of \textit{valid} candidates $\Omega$ (see \autoref{fig:graph_and_candidate}, right).
Specifically, valid candidate start-nodes must have an out-degree smaller than four whereas candidate end-nodes must have an in-degree less than or equal to one.
We also exclude the starting set of game materials from the candidate end-nodes.
We call this technique \emph{candidate sampling}.

Afterwards, a decoder classifies edges as positive or negative by relying on the node embeddings. 
In particular, we learn a scoring function $s: \mathbb{R}^d \times \mathbb{R}^d \to \mathbb{R}$ using a linear layer $f$ that takes as input the concatenation of the two node embeddings corresponding to the candidate edge, and the output $c$ of the sequence model, that serves as a context: $s(v_i, v_j) = \hat{y}_{ij} = f(z_i \oplus z_j \oplus c)$,
\begin{table}[t]
  \centering
  \resizebox{\linewidth}{!}{
  \begin{tabular}{lccc}
    \toprule
    \multicolumn{4}{c}{\textit{Status}} \\
    \midrule
    Modalities & \citet{bara2023towards} & Ours & Human\\
    \midrule
    M  & $47.7\pm0.6$ & $59.9\pm0.7$ & $67.0$ \\
    D+M & $45.5\pm2.3$ & $59.1\pm0.6$ & $67.0$ \\
    D+V+M & $45.2\pm1.8$ & $58.9\pm0.8$ & $67.0$ \\
    V+M & $47.3\pm0.7$ & $59.6\pm0.4$ & $67.0$ \\
    \midrule
    \multicolumn{4}{c}{\textit{Knowledge}} \\
    \midrule
    Modalities & \citet{bara2023towards} & Ours & Human\\
    \midrule
    M  & $51.5\pm1.1$ & \cellcolor{green!25} $57.9\pm0.2$ & \cellcolor{green!25} $58.0$ \\
    D+M & $50.0\pm1.5$ & \cellcolor{green!25} $57.2\pm1.5$ & \cellcolor{green!25} $58.0$ \\ 
    D+V+M & $50.2\pm1.1$ & \cellcolor{green!25} $57.5\pm1.7$ & \cellcolor{green!25} $58.0$ \\
    V+M & $50.5\pm1.6$ & \cellcolor{green!25} $57.6\pm1.8$ & \cellcolor{green!25} $58.0$ \\
    \midrule
    \multicolumn{4}{c}{\textit{Intention}} \\
    \midrule
    Modalities & \citet{bara2023towards} & Ours & Human\\
    \midrule
    M  & $9.1\pm0.2$ & $11.7\pm2.2$ & $46.0$ \\
    D+M & $8.7\pm2.1$ & $11.1\pm1.8$ & $46.0$ \\
    D+V+M & $10.5\pm2.3$ & $12.1\pm2.4$ & $46.0$ \\
    V+M & $9.0\pm0.3$ & $13.4\pm1.9$ & $46.0$ \\
    \bottomrule
  \end{tabular}
  }
  \caption{Performance comparison on the three ToM tasks using different combinations of modalities: dialogue moves (M), dialogue (D), and video frames (V). We report the F1 scores obtained by the baseline~\cite{bara2023towards}, our model, and humans.
  }
  \label{tab:tom_tasks}
\end{table}
\noindent where $\oplus$ denotes the concatenation operator.
In contrast to \cite{bara2023towards} who used an LSTM to process the sequential data, we used a single Transformer block.
Finally, we optimise with the binary cross-entropy loss function $\mathcal{L}$, which maximises the likelihood of positive edges while minimising that of  sampled negative edges:
\begin{equation*}
    \mathcal{L} = - \sum_{(i, j) \in E} \log(\sigma(\hat{y}_{ij})) - \sum_{(i, k) \in \Omega} \log(1 - \sigma(\hat{y}_{ik}))
\end{equation*}
where $\sigma$ indicates the sigmoid function. 
Additional details about the model's architecture and training are provided in \S\ref{app:model}.

%% file: latex/04_experiments.tex
\begin{table*}[t]
    \centering
    \resizebox{\linewidth}{!}{
    \begin{tabular}{cccccccccc}
    \toprule
    \multicolumn{3}{c}{ToM Features} & \multicolumn{2}{c}{Overall} & \multicolumn{3}{c}{OMK} & \multicolumn{2}{c}{PMK} \\
    \cmidrule(lr{0.05cm}){1-3} \cmidrule(lr{0.05cm}){4-5} \cmidrule(lr{0.05cm}){6-8} \cmidrule(lr{0.05cm}){9-10}
    \textit{Status} & \textit{Knowledge} & \textit{Intention} & \citet{bara2023towards} & Ours & 
     \citet{bara2023towards} & Ours (NS) & Ours & \citet{bara2023towards} & Ours \\
    \midrule
    & & & $46.6 \pm 1.6$ & $\mathbf{56.9} \pm 0.6$ & $27.7 \pm 2.3$ & \cellcolor{cyan!25} $23.7\pm2.5$ & \cellcolor{cyan!25} $57.6\pm 0.8$ & $65.4\pm 0.2$ & $56.2\pm 0.3$ \\
    \checkmark & & & $46.7 \pm 2.0$ & $\mathbf{57.3}\pm 0.6$ & $26.1 \pm 2.5$ & \cellcolor{cyan!25} $26.6\pm 2.4$ & \cellcolor{cyan!25} $58.0\pm 0.8$ & $67.2\pm 1.2$ & $56.5\pm 0.3$ \\
    & \checkmark & & $47.4\pm 1.7$ & $\mathbf{57.0}\pm 1.4$ & $28.0 \pm 1.8$ &\cellcolor{cyan!25} $24.7\pm 2.6$ & \cellcolor{cyan!25} $58.4\pm 0.5$ & $66.8\pm 1.5$ & $55.5\pm 1.9$ \\
    & & \checkmark & $47.2\pm 1.9$ & $\mathbf{57.2}\pm 0.5$ & $28.0 \pm 2.6$ & \cellcolor{cyan!25} $26.0\pm 1.2$ & \cellcolor{cyan!25} $57.9\pm 0.7$ & $66.3\pm 0.8$ & $56.5\pm 0.3$ \\
    \checkmark & \checkmark & & $47.6\pm 1.5$ & $\mathbf{56.6}\pm 1.4$ & $28.4 \pm 1.4$ & \cellcolor{cyan!25} $25.2\pm 0.3$ & \cellcolor{cyan!25} $57.7\pm 0.5$ & $66.8\pm 1.5$ & $55.5\pm 1.9$ \\
    \checkmark & & \checkmark & $47.6\pm1.7$ & $\mathbf{57.5}\pm0.6$ & $28.4 \pm 1.8$ & \cellcolor{cyan!25} $27.2\pm 1.3$ & \cellcolor{cyan!25} $58.4\pm 0.8$ & $66.8\pm 1.5$ & $56.5\pm 0.3$ \\
    & \checkmark & \checkmark & $47.2\pm1.7$ & $\mathbf{57.5}\pm0.6$ & $27.6 \pm 1.9$ & \cellcolor{cyan!25} $27.7\pm 0.7$ & \cellcolor{cyan!25} $58.5\pm 0.8$ & $66.8\pm 1.5$ & $56.4\pm 0.1$ \\
    \checkmark & \checkmark & \checkmark & $47.4\pm1.8$ & $\mathbf{56.7}\pm0.7$ & $27.9 \pm 2.0$ & \cellcolor{cyan!25} $26.6\pm 2.9$ & \cellcolor{cyan!25} $57.1\pm 1.9$ & $66.8\pm 1.5$ & $56.6\pm 0.2$ \\
    \bottomrule
    \end{tabular}
    }
    \caption{Performance comparison on CPA when training with learnt ToM features. We report the overall F1 scores as well those for own (OMK) and partner (PMK)  missing knowledge prediction. NS = Naive Sampling.}
    \label{tab:cpa_tom_feats}
\end{table*}

\begin{table}[t]
    \centering
    \begin{tabular}{lcccc}
        \toprule
        ToM Task & ToM & OMK & PMK & Random\\
        \midrule
        \textit{Status} & $60.6$ & $51.6$ & $49.5$ & $46.7$ \\
        \textit{Knowledge} & $50.9$ & $49.8$ & $50.8$ & $45.1$ \\
        \textit{Intention} & $10.2$ & $14.1$ & $13.0$ & $9.3$ \\
        \bottomrule
    \end{tabular}
    \caption{F1 scores on ToM tasks for logistic regression models trained using ToM features, CPA features and random noise. OWM and PMK indicate features coming from our model trained, without ToM features as input, on one's own and partner's missing knowledge prediction, respectively. \vspace{-0.3cm}}
    \label{tab:logreg}
\end{table}

\section{Experiments}
\label{sec:experiments}

\subsection{Theory of Mind Modelling}

We first report the performance of our model on the three ToM tasks introduced in Section \ref{sec:prob_form}. 
As summarised in Table~\ref{tab:tom_tasks}, our model outperforms the baseline\footnote{Despite training the baseline model~\cite{bara2023towards} using the \href{https://github.com/sled-group/collab-plan-acquisition}{official code}, its performance slightly deviated from the original paper, and discussions with the authors did not yield clarity. See \S\ref{app:bara} for further details and comparisons.} on all three tasks, underlining the efficiency of the proposed GNN-based approach. 
Notably, as highlighted in \colorbox{green!25}{green} in \autoref{tab:tom_tasks}, our model manages to even match human performance in the \textit{Knowledge} task. 
However, performance on the other two tasks is still far from a human level, especially on \textit{Intention}. 
This might be attributed to the fact that, unlike \textit{Knowledge} that does not require temporal modelling and could be solved by using plan information,
both \textit{Status} and \textit{Intention} require accurate temporal modelling, which has to be kept coherent across the different input modalities. 
As can also be seen from the table, ablations of different input modalities have little impact on the final performance of our model and the baseline for all ToM tasks.  

\subsection{Collaborative Plan Acquisition (CPA)}
\label{subsec:cpa}
Subsequently, we evaluate our model on the CPA task following \citet{bara2023towards}. We always use dialogue moves as input since they were shown to have 
a positive impact on performance. 
As can be seen from \autoref{tab:cpa_tom_feats}, our model consistently achieves overall F1 scores of over $56.6$ thereby significantly outperforming the baseline of \citet{bara2023towards} in all evaluation settings. 
\paragraph{Own Missing Knowledge (OMK).}
We first analyse the task of predicting one's own missing knowledge.
As can be seen in \autoref{tab:cpa_tom_feats}, our model manages to double the performance of the baseline\footnote{In this case, the scores are higher than the ones reported in \cite{bara2023towards}.} by consistently achieving F1 scores of over $57\%$.
In stark contrast to the baseline, which performs best when using only ToM features extracted from \textit{Intention}, our model's best performance is obtained by additionally incorporating features extracted from \textit{Knowledge}. 
The benefit of these features on CPA 
is expected and intuitively makes sense since \textit{Knowledge} was the ToM task for which 
our models achieved human-level performance (see \autoref{tab:tom_tasks}). 

\paragraph{Partner's Missing Knowledge (PMK).} 
Second, we evaluate the task of predicting the partner's missing knowledge.
In contrast to prior work~\citep{bara2023towards}, our evaluations reveal a significantly reduced performance gap between predicting the different types of missing knowledge (OMK vs PMK) as can be seen in the second part of \autoref{tab:cpa_tom_feats}.
As highlighted in \colorbox{cyan!25}{blue}, this can be attributed to our proposed candidate sampling approach that, contrarily to naive sampling, effectively narrowed down the pool of valid candidate edges for one's own missing knowledge to a similar order of magnitude as that of the partner's.
The difference in performance compared to the baseline is likely due to the choice of cost function used for training.

\paragraph{Statistical Tests.}
Although our model attained improved results on CPA, especially in predicting one's own missing knowledge, it did so without relying much on the learned ToM features.
This can be seen from the results of the different ablated versions of \autoref{tab:cpa_tom_feats}. 
To study the effect of ToM features on CPA performance, we performed paired t-tests between our model trained without ToM features and versions of our model trained with different sets of ToM features. 
We can see that the ToM features did not result in any statistically significant performance difference on CPA since all tests resulted in $p > 0.05$.
Notably, performing the significance testing on the baseline model of \cite{bara2023towards} yielded the same behaviour, i.e. $p>0.05$ across all model versions.
Therefore, we challenge the utility of the ToM features in CPA by posing the question of {whether these features represent actual information about mental states or reflect latent patterns in the data}. 
We empirically answer this by performing various principled experiments ranging from diagnostic probing to correlation analysis over substituting ToM features with ground-truth labels.

\subsection{Probing for Theory of Mind}

Motivated by our results, we formulate the following research question: \textit{Does ToM modelling as proposed by \citet{bara2023towards} actually  capture mental state information?}
To answer this question, we conducted extensive analyses to study the impact of ToM modelling on CPA from different angles.

\subsubsection{Diagnostic Probing}

The ToM features used in CPA are obtained by learning the different tasks of Section \ref{sec:prob_form}.  
As a result, we expect such features to exclusively hold some information about the mental state that other models, when trained on different tasks than ToM, simply lack.    
To validate this intuitive hypothesis, we used diagnostic probing~\citep{alain2016understanding, adi2017finegrained, conneau-etal-2018-cram, hupkes2018visualisation} and
trained a simple logistic regression (LR) model to perform the three ToM classification tasks.
We trained the LR model with different inputs in each experiment and tested its performance on the ToM tasks using the test split.
More specifically, we considered four different input scenarios: the vanilla ToM features used in the previous experiments, the hidden representations of the transformer from models predicting OMK and PMK (output of the green model in \autoref{fig:model}), and finally random noise. 
As seen in \autoref{tab:logreg}, a LR model trained on ToM features can perform reasonably well on the three tasks, 
especially \textit{Status}. 
However, when trained with missing knowledge features, i.e. features completely optimised in the absence of ToM, the LR model achieves comparable performance in \textit{Knowledge} and even better performance in \textit{Intention}.  

These findings open up two possible scenarios: (1) The learnt ToM features are more likely to represent latent patterns in the data with no perceivable link to ToM; 
(2) ToM capabilities spontaneously emerge from training models on CPA.
\begin{figure*}[t]
    \centering
    \includegraphics[width=\linewidth]{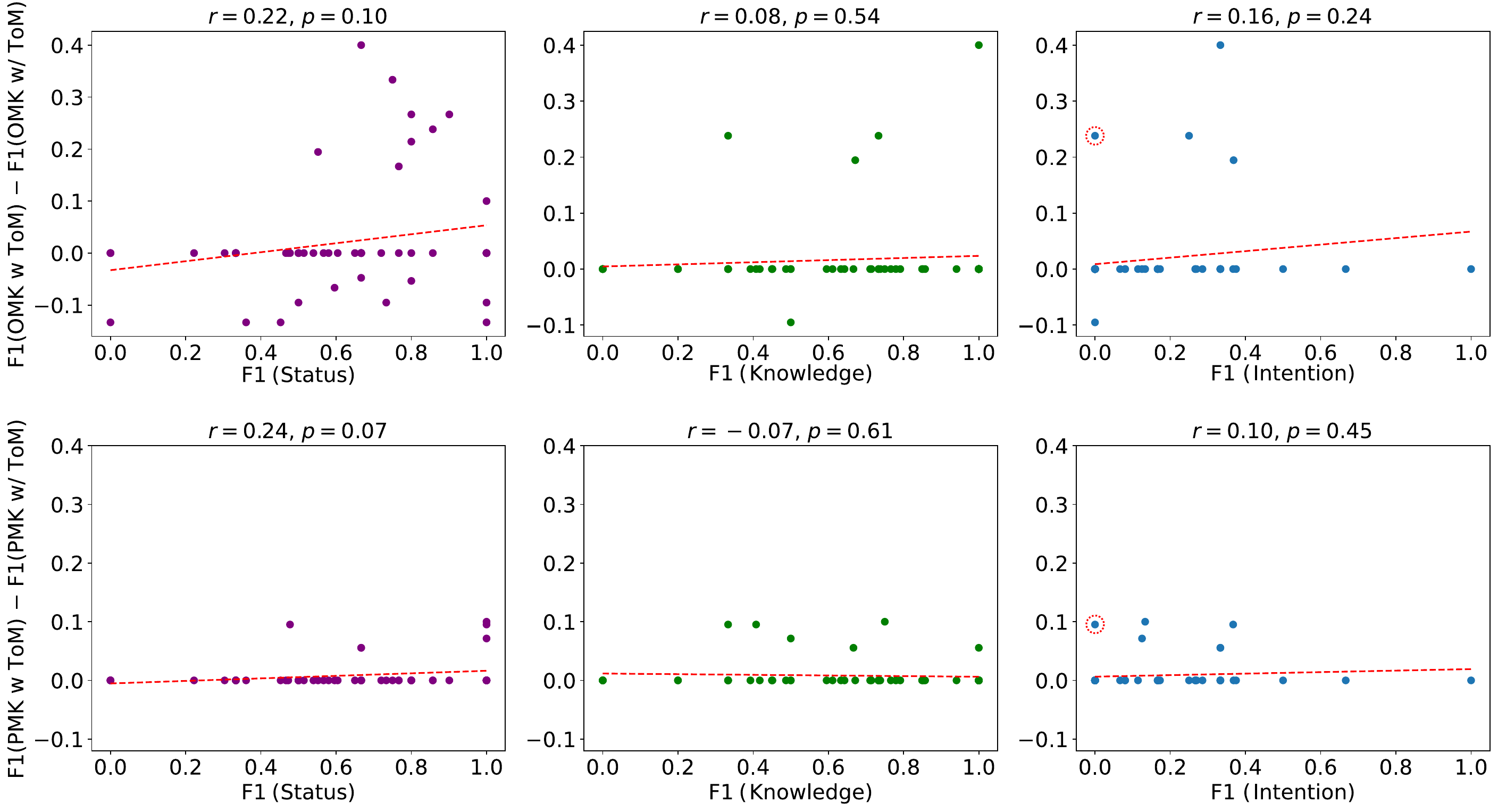}
    \caption{Correlation between F1 scores in the ToM tasks and the difference between F1 scores on OMK \textbf{(top)} and PMK \textbf{(bottom)} obtained by our model trained with and without ToM features as input. The red dotted line represents a linear fit. The red dotted circle indicates game sessions for which performance on CPA improved even if the F1 score on ToM task is zero.
    }
    \label{fig:correlation}
\end{figure*}
\subsubsection{Correlation Analysis} 

In this experiment, we explored whether improvements in CPA tasks correlate with the performance of models on ToM tasks, which are then used to extract ToM features. 
Intuitively, if the ToM features truly benefit the model, we anticipate a strong positive correlation between the performance on the individual task ToM tasks and that of CPA. 
To validate this, we chose the best-performing model for each 
ToM feature in CPA and calculated the performance difference relative to the model without any ToM features (refer to \autoref{tab:cpa_tom_feats}).
Next, we calculated the Pearson correlation coefficient to measure the relationship between this difference in performance and the F1 score on ToM tasks.
Results for both OMK and PMK are reported in \autoref{fig:correlation}.
Given that $r < 0.3$ and $p \gg 0.05$ in all cases, we can conclude that \emph{there is no correlation between the performance on ToM tasks and improvements on CPA}. 
It is worth noting that for some cases,
CPA improved even if the F1 score on ToM was zero (highlighted with red dotted circles in \autoref{fig:correlation}). 

\begin{table*}[t]
    \centering
    \resizebox{\linewidth}{!}{
    \begin{tabular}{ccccccc}
    \toprule
    \multicolumn{3}{c}{ToM Labels} & \multicolumn{2}{c}{OMK} & \multicolumn{2}{c}{PMK} \\
    \cmidrule(lr{0.05cm}){1-3} \cmidrule(lr{0.05cm}){4-5} \cmidrule(lr{0.05cm}){6-7}
    \textit{Status} & \textit{Knowledge} & \textit{Intention} 
    & \citet{bara2023towards} & Ours & \citet{bara2023towards} & Ours \\
    \cmidrule(lr{0.05cm}){1-3} \cmidrule(lr{0.05cm}){4-5} \cmidrule(lr{0.05cm}){6-7}
    & & & $26.3\pm 1.9$ & $58.2\pm 0.3$ & $60.9\pm 3.2$ & $51.5\pm 4.7$ \\
    \checkmark & & & $26.8\pm 1.6$ & $58.5\pm 0.6$ & $66.0\pm 1.9$ & $51.5\pm 4.7$ \\
    & \checkmark & & $26.8\pm 1.6$ & $58.3\pm 0.2$ & $66.0\pm 1.9$ & $51.5\pm 4.7$ \\
    & & \checkmark & $26.8\pm 1.6$ & $58.2\pm 0.3$ & $66.0\pm 1.9$ & $52.2\pm 3.4$ \\
    \checkmark & \checkmark & & $26.6\pm 1.2$ & $58.3\pm 0.2$ & $66.0\pm 1.9$ & $51.5\pm 4.7$ \\
    \checkmark & & \checkmark & $27.0\pm 1.4$ & $58.4\pm 0.2$ & $66.0\pm 1.9$ & $51.5\pm 4.7$ \\
    & \checkmark & \checkmark & $26.9\pm 1.6$ & $58.6\pm 0.5$ & $66.0\pm 1.9$ & $51.0\pm 4.2$ \\
    \checkmark & \checkmark & \checkmark & $26.6\pm 1.1$ & $58.5\pm 1.3$ & $66.0\pm 1.9$ & $51.5\pm 4.7$ \\
    \bottomrule
    \end{tabular}
    }
    \caption{Performance comparison on CPA when training with ground-truth ToM labels.
    We report F1 scores for own (OMK) and partner (PMK) missing knowledge prediction.
    }
    \label{tab:gt_tom}
\end{table*}

\subsubsection{Incorporating ToM Ground-Truth}

In our final experiment, we aim to assess the utility of mental state information in CPA by answering the question: \textit{To what extent do models gain from including ground-truth information about mental states?} 
We replicate our experiments from Section \ref{subsec:cpa} to explore this.
However, instead of the learnt ToM features, we feed models with 
a one-hot encoding of the corresponding ToM question-answer pair.
This encoding process follows the same methodology employed by \citet{bara-etal-2021-mindcraft}.
As can be seen from a comparison of \autoref{tab:gt_tom} and \autoref{tab:cpa_tom_feats}, the baseline and our model trained with 
ground-truth mental state information consistently under-perform those trained with learnt ToM features on OMK and PMK, respectively.
The remaining results are generally on par.
Furthermore, upon comparing the initial row in  \autoref{tab:gt_tom} with the subsequent rows, we note that incorporating ToM ground-truth yields similar scores to those achieved without, except for the baseline in PMK.
This underscores a significant limitation of the ToM task representation: \textit{the collected ground-truth mental states are not beneficial for CPA}.
This finding, in conjunction with our diagnostic probing analysis, suggests that models trained to infer mental states may be learning information more closely associated with other correlations in the data, rather than representing the mental states.

%% file: latex/05_conclusion.tex
\section{Limits and Future Directions for Neural Theory of Mind} 

A key insight of our work is that current approaches for CPA, rather than learning ToM, seem to merely exploit latent correlations in the data that have little to do with mental states.
This is highlighted by the lack of impact of the proposed ToM features on CPA, as shown in \autoref{tab:cpa_tom_feats} and \autoref{tab:gt_tom}.
This finding is surprising and worrisome at the same time and calls for a fundamental re-assessment of how to equip computational agents with ToM capabilities and how to evaluate them.
Despite research on this topic still being in its infancy, 
the problem of correctly learning neural ToM has recently been put more and more under scrutiny~\citep{sap-etal-2022-neural, Aru2023Mind}.
Our results underline in a directly observable way that the acquisition of comprehensive ToM capabilities cannot be reduced to merely passing a specific, narrow set of tasks. 
The main rationale for this conclusion is that we still do not have a task for which possessing ToM capabilities is both a \textit{necessary and sufficient} prerequisite for its resolution. 
Current ToM benchmarks rely on tasks that seem to \textit{intuitively} require ToM to be solved.
However, these tasks can often be solved by just exploiting shortcuts within the data~\cite{le-etal-2019-revisiting,Aru2023Mind,bortoletto24_aaai}.
\textbf{As a result, we posit that directly optimising an agent or system for ToM may not represent an effective approach for progress}.

Instead, recent work proposed the use of open-ended environments to study ToM with the aim of observing whether these capabilities emerge through interactions with other agents~\citep{Aru2023Mind}. 
Minecraft represents a good candidate environment for multi-agent collaboration in an open world.
However, the way \citet{bara-etal-2021-mindcraft, bara2023towards} frame \textit{MindCraft} is still limited to specific tasks and requires extensive data collection efforts.
One possible solution could be to transform \textit{MindCraft} into a reinforcement learning environment with a focus on less constrained collaborative tasks.
While \citet{bara-etal-2021-mindcraft, bara2023towards} suggest modelling ToM as a supervised learning task, the way humans acquire ToM is more nuanced and largely unsupervised~\citep{ruffman2023belief}.
\textbf{We believe that the development of open-ended environments combined with learning ToM capabilities in an unsupervised, human-like manner is a more promising direction for future research}. 

ToM capabilities are deeply linked to language acquisition~\citep{tomasello2005constructing}.
In the context of dialogue-based collaboration, another interesting future direction could be to learn ToM from generation instead of classification~\citep{liu2023computational}.
Current approaches could be further improved by building a more general and robust world model, e.g., by leveraging a pre-trained language or video-language model as a more general prior. 
Finally, in addition to developing suitable environments and learning algorithms, effective and interpretable methods to evaluate whether agents have truly learned ToM will be crucial. 
We see three exciting directions in this regard: probing \citep{Niven2019Probing}, mechanistic interpretability \citep{wang2022interpretability}, and concept learning \citep{Oguntola2021Deep,Chen2020Concept}.
The work of \citet{oguntola2023theory} serves as an inspiring example, where agents learn human-interpretable concepts that represent beliefs about other agents in a simple multi-agent reinforcement learning setting.

\section{Conclusion}
\label{sec:conclusion}

In this work, we demonstrated that applying task-specific constraints to plan graphs reduces significantly the performance gap between predicting OMK and PMK in \textit{MindCraft}.
At the same time, improvements from ToM modelling diminish, raising concerns about current approaches.
Our experiments and analyses consistently suggest that current ToM modelling approaches learn features that likely reflect latent patterns in the data, with no perceivable link to ToM.
This finding calls for a deeper understanding of the role of ToM in CPA and beyond, as well as for new methods to model and evaluate mental states in collaborative agents.

%% file: latex/06_limitations.tex
\section{Limitations}
\label{sec:limitations}

We identify two main limitations of our work.
First, our strategy of selecting candidate edges for own missing knowledge prediction is specific to the structure of the task as presented by \citet{bara-etal-2021-mindcraft}.
While we recognise the task-specific nature of our strategy, it is crucial to note that \citet{bara2023towards} also leverages task-specific constraints by assuming that the partner's missing knowledge is present in one's own.
However, our approach can still be used in other settings without the assumption that one's own missing knowledge covers that of the partner. 
Crucially, our approach does not challenge or undermine the fundamental conclusions drawn about modelling and evaluating ToM. Neither does it serve as the cause for the diminishing improvements on CPA observed when including ToM features.

Second, our current analysis is limited to the \textit{MindCraft} dataset. 
To the best of our knowledge, it is the only environment that studies the role of ToM in CPA making it the natural candidate for our analysis.
However, our work lays the basis of a systematic study of ToM in CPA in general and can also inform future work targeting new environments or datasets.

%% file: latex/07_ethics.tex
\section{Ethical Impact}
\label{sec:ethics}
Our work is foundational, far away from particular applications or any potential societal impact. 
However, it is important to keep in mind that claims about modelling and predicting mental states potentially have huge ethical impact. 
Caution is imperative when dealing with sensitive aspects of individuals' inner experiences and emotions. 
Mishandling such information could lead to privacy breaches, potential stigmatisation, or the misuse of personal data. 
Additionally, there is a risk of reinforcing biases or misinterpreting complex psychological nuances, which may have unintended consequences on individuals' well-being. 
Lastly, resonating with our findings, the use of models that predict mental states by merely exploiting heuristics and spurious patterns in the data rather than genuinely modelling Theory of Mind introduces significant ethical challenges. 
Therefore, ethical considerations and responsible practices are crucial to ensure a respectful and appropriate use of technology in this domain.

\section{Acknowledgements}
M. Bortoletto and A. Bulling were funded by the European Research Council (ERC) under the European Union’s Horizon 2020 research and innovation programme under grant agreement No 801708.
L. Shi was funded by the Deutsche Forschungsgemeinschaft (DFG, German Research Foundation) under Germany’s Excellence Strategy -- EXC 2075 -- 390740016.
The authors thank the International Max Planck Research School for Intelligent Systems (IMPRS-IS) for supporting C. Ruhdorfer.
The authors would like to especially thank Hsiu-Yu Yang, Manuel Mager, Pavel Denisov, Ekta Sood, and Anna Penzkofer for their support and the numerous insightful discussions.

%% file: latex/08_appendix.tex
\section{Appendix}

\subsection{MindCraft}
\label{app:mc1}
MindCraft~\citep{bara-etal-2021-mindcraft} is an interactive environment based on Minecraft in which the objective of each game is to create a randomly generated goal material. 
Some starting materials are already present in the environment, and the others are created by the players by: 
\begin{itemize}
    \item \textit{Mining}: hit a specific block with a tool to create a new block. This allows to generate infinite new blocks of a certain type and move them around.
    \item \textit{Combining}: fuse two materials to obtain a new one. 
\end{itemize} 
Players are free to navigate in the environment, move blocks around and chat. Being the field of view first-person, players have partial observability of the environment.  

Players possess \textit{asymmetric knowledge and skill sets}, where each player is provided with a partial plan (such as a partial knowledge graph or recipe) along with a specific tool to interact with certain blocks. 
Their knowledge and skills are complementary: Player 1's plan contains the information missing in Player 2's plan, and vice versa. 
Likewise, Player 1 can interact with blocks that Player 2 cannot, and vice versa. 
This inherent asymmetry encourages communication between players and the need to reason about each other's mental states.

\citet{bara-etal-2021-mindcraft} introduced three ToM question answering tasks that are specifically designed to capture mental state information that is pertinent to collaboration. 
Players are presented with three questions, each paired by type: if one player is asked about their partner's beliefs, the other player is presented with the same question regarding their own beliefs (with respect to the same question):
\begin{enumerate}[leftmargin=*, itemsep=0pt]
    \item \textit{Task Status}: Predict if a specific material has been created by one of the two players. 
    For example, if Player 1 is asked: \texttt{Has the other player made GOLD\_BLOCK until now?}, then Player 2 is asked: \texttt{Have you created GOLD\_BLOCK until now?}. 
    Possible answers are \texttt{YES}, \texttt{NO}, or \texttt{MAYBE}. 
    This task probes players' belief of the game state.
    \item \textit{Player Knowledge}: Predict whether a player knows how to create a specific material or if they believe their partner knows. 
    For example, if Player 1 is asked: \texttt{Do you think the other player knows how to make BLUE\_WOOL?}, then Player 2 is asked: \texttt{Do you know how to make BLUE\_WOOL?}.
    Possible answers are \texttt{YES}, \texttt{NO}, or \texttt{MAYBE}. 
    This task probes players' belief of their and their partner's current knowledge.
    \item \textit{Player Intention}: Predict which material a player is making at the current time step. 
    For example, if Player 1 is asked: \texttt{What do you think the other player is making right now?}, then Player 2 is asked: \texttt{What are you making right now?}. 
    Possible answers are the different types of block in the game or \texttt{NOT\_SURE}. 
\end{enumerate}
Mental state annotations are collected using periodic pop-ups that interrupt the game every 75 seconds. 

Computational models are trained and evaluated by predicting the answer to the ToM tasks from a player's perspective, given a history of observations, the chat dialogue, the perceived actions in the shared environment, and the partial plan. 
Models are trained to minimise the cross entropy loss.
They output $c$ logits, with $c$ being the number of possible classes: three for \textit{Status} and \textit{Knowledge}, and 22 for \textit{Intention} (21 possible materials + \texttt{NOT\_SURE}). 
Models' final prediction is obtained by taking the argmax of the logits.

\subsubsection{Data Modalities}
The \textit{MindCraft} dataset comprises various modalities, including first-person video streams for each player, chat dialogue and corresponding dialogue moves, and players' plans. 
\citet{bara-etal-2021-mindcraft} established a timestep of $\Delta t = 1$ second, ensuring that each timestep has an associated video frame. 
However, not all timesteps include dialogue utterances or questions, as players do not necessarily exchange messages every second and questions pop-up every 75 seconds. 
The players' plan is static and known from the beginning of the game.
The models generate predictions at timestep $t$, coinciding with when questions are posed to the players. 
These predictions rely on the player's plan, video stream, and dialogue history (if available) up to time $t$.

\subsubsection{Dataset Statistics}
\citet{bara-etal-2021-mindcraft} report that the original MindCraft dataset includes 100 games, with an average of 20.5 dialogue exchanges per game, for a total of 2091 exchanges. 
Games last between 1 minute and 22 seconds to 27 minutes and 26 seconds, with the average game lasting 7 minutes and 23 seconds. 
A total of 12 hours, 18 minutes, and 33 seconds of in-game interaction was recorded. 
Between 5 and 10 objects are used in a game, and between 7 and 11 steps are necessary in each game to achieve the goal. 
The dataset is randomly partitioned into 60\% for training, 20\% for validation, and 20\% for testing, with the condition to keep similar distributions of game lengths.

\subsection{Collaborative Plan Acquisition}
\label{app:mc2}
\citet{bara2023towards} extended MindCraft by collecting 60 additional game sessions and defined an additional task: \textit{collaborative plan acquisition} (CPA).
In CPA a model has to predict, from a player's perspective, its own missing knowledge (OMK) and the other player's missing knowledge (PMK). 
CPA was introduced to explore the role of ToM modelling in predicting players' missing knowledge while executing the crafting tasks.  
CPA is formulated as follows: 

\medskip
\noindent\textbf{Definition 1} \textit{
    Consider a joint plan as a directed AND-graph $\mathcal{P} = (V,E)$, where the nodes $V$ denote (sub-)goal materials, and edges $E$ denote temporal constraints between the sub-goals. In a collaborative plan acquisition problem, two agents $i$ and $j$ start with partial plans $\mathcal{P}_i = (V, E_i)$, $E_i \subseteq E$, and $\mathcal{P}_j = (V, E_j)$, $E_j \subseteq E$. Given a sequence of visual observations $O^t_i$ and a joint dialogue history $D^t$ at time $t$, agent $i$ has to infer their own missing knowledge $\Bar{E}_i = E \setminus E_i$ and the partner $j$’s missing knowledge $\Bar{E}_j = E \setminus E_j$.
}

\medskip
Predictions for OMK and PMK are made at $t=T$, where $T$ is the final timestep of the game. 

While intrinsically linked, ToM and CPA tasks exhibit a fundamental distinction: ToM tasks directly explore players' beliefs regarding the game state and their partner's mental states, whereas CPA tasks involve predicting the absent information from players' partial plans. 
In ToM tasks, the ground truth is based on players' beliefs, which may be true or false, whereas in CPA tasks, the ground truth is formally determined by $\Bar{E}_i = E \setminus E_i$ for OMK and $\Bar{E}_j = E \setminus E_j$ for PMK. 
\citet{bara2023towards} discuss the partial overlap between \textit{Task Knowledge} and PMK, highlighting their difference: \textit{Task Knowledge} probes whether a \textit{single} piece of knowledge is known by the partner, while PMK involves  predicting whether the partner shares \textit{each} piece of the player's knowledge.

\subsubsection{Data Modalities}
CPA is conducted in the MindCraft environment, therefore the data modalities utilised mirror those in the ToM tasks: first-person video streams for each player, chat dialogue along with corresponding dialogue moves, and the players' plans. 
In addition to the aforementioned modalities, models for CPA receive ToM features extracted from the sequence-to-sequence model trained on the ToM tasks discussed in \S\ref{app:mc1}. 
These ToM features are tensor representations of dimension $1024$ that are incorporated into the model's input.
A timestep of $\Delta t = 1$ second is maintained.
In contrast to the ToM tasks, CPA models generate predictions at timestep $t=T$, i.e., at the end of each game.

\subsubsection{Formalising ToM Tasks}
Based on the formalism introduced for CPA, we can formalise the ToM tasks as follows:
\begin{enumerate}[leftmargin=*, itemsep=0pt]
    \item \textit{Task Status}: Predict if a specific material $V_k \in V$ has been created by one of the two players. 
    \item \textit{Player Knowledge}: Predict whether a player knows how to create a specific material $V_k$, i.e., 
    \begin{align*}
        \{e_{V_k}^n\}_{n>1} \stackrel{\tiny ?}{\in} E_i
    \end{align*}
    or if they believe their partner knows, i.e.\, 
    \begin{align*}
        \{e_{V_k}^n\}_{n>1} \stackrel{\tiny ?}{\in} E_j,
    \end{align*}
    with $e_{V_k}\in E$ being an edge with end-node $V_k$ and $n$ being the number of materials needed to craft $V_k$. 
    \item \textit{Player Intention}: Predict which material $V_k \in V$ a player is making at the current time step $t$. 
\end{enumerate}

\subsection{Technical Details}
\label{app:model}

\begin{table}[t]
  \centering
  \resizebox{\linewidth}{!}{
  \begin{tabular}{lccc}
    \toprule
    \multicolumn{4}{c}{\textit{Status}} \\
    \midrule
    Modalities & \citet{bara2023towards} & Ours & Human\\
    \midrule
    M  & $56.0\pm0.8$ & $59.9\pm0.7$ & $67.0$ \\
    D+M & $54.6\pm1.1$ & $59.1\pm0.6$ & $67.0$ \\
    D+V+M & $59.3\pm1.0$ & $58.9\pm0.8$ & $67.0$ \\
    V+M & $59.3\pm1.7$ & $59.6\pm0.4$ & $67.0$ \\
    \midrule
    \multicolumn{4}{c}{\textit{Knowledge}} \\
    \midrule
    Modalities & \citet{bara2023towards} & Ours & Human\\
    \midrule
    M  & $54.7\pm2.5$ & $57.9\pm0.2$ & $58.0$ \\
    D+M & $56.2\pm1.9$ & $57.2\pm1.5$ & $58.0$ \\ 
    D+V+M & $57.6\pm1.0$ & $57.5\pm1.7$ & $58.0$ \\
    V+M & $56.4\pm2.5$ & $57.6\pm1.8$ & $58.0$ \\
    \midrule
    \multicolumn{4}{c}{\textit{Intention}} \\
    \midrule
    Modalities & \citet{bara2023towards} & Ours & Human\\
    \midrule
    M  & $14.9\pm0.2$ & $11.7\pm2.2$ & $46.0$ \\
    D+M & $12.1\pm1.0$ & $11.1\pm1.8$ & $46.0$ \\
    D+V+M & $13.5\pm0.6$ & $12.1\pm2.4$ & $46.0$ \\
    V+M & $13.8\pm1.7$ & $13.4\pm1.9$ & $46.0$ \\
    \bottomrule
  \end{tabular}
  }
  \caption{Performance comparison on the three ToM tasks using different combinations of modalities: dialogue moves (M), dialogue (D), and video frames (V). F1 scores for the baseline are reported from the original paper~\cite{bara2023towards} rather than from our execution of the official code.}
  \label{app:tab:tom_tasks}
\end{table}

\begin{table*}[t]
    \centering
    \resizebox{\linewidth}{!}{
    \begin{tabular}{ccccccccc}
    \toprule
    \multicolumn{3}{c}{ToM Features} & \multicolumn{2}{c}{Overall} & \multicolumn{2}{c}{OMK} & \multicolumn{2}{c}{PMK} \\
    \cmidrule(lr{0.05cm}){1-3} \cmidrule(lr{0.05cm}){4-5} \cmidrule(lr{0.05cm}){6-9} 
    \textit{Status} & \textit{Knowledge} & \textit{Intention} & \citet{bara2023towards} & Ours & 
     \citet{bara2023towards} & Ours & \citet{bara2023towards} & Ours \\
    \midrule
    & &                                  & $44.1 \pm 0.6$ & $\mathbf{56.9} \pm 0.6$ & $16.7 \pm 0.1$ & $57.6 \pm 0.8$ & $71.4 \pm 1.0$ & $56.2 \pm 0.3$ \\
    \checkmark & &                       & $45.9 \pm 1.5$ & $\mathbf{57.3} \pm 0.6$ & $20.4 \pm 1.4$ & $58.0 \pm 0.8$ & $71.3 \pm 1.6$ & $56.5 \pm 0.3$ \\
    & \checkmark &                       & $47.2 \pm 1.1$ & $\mathbf{57.0} \pm 1.4$ & $20.1 \pm 1.4$ & $58.4 \pm 0.5$ & $74.3 \pm 0.7$ & $55.5 \pm 1.9$ \\
    & & \checkmark                       & $47.4 \pm 1.4$ & $\mathbf{57.2} \pm 0.5$ & $19.8 \pm 1.7$ & $57.9 \pm 0.7$ & $75.0 \pm 1.0$ & $56.5 \pm 0.3$ \\
    \checkmark & \checkmark &            & $47.0 \pm 1.4$ & $\mathbf{56.6} \pm 1.4$ & $20.9 \pm 1.2$ & $57.7 \pm 0.5$ & $73.1 \pm 1.5$ & $55.5 \pm 1.9$ \\
    \checkmark & & \checkmark            & $45.9 \pm 1.2$ & $\mathbf{57.5} \pm 0.6$ & $19.8 \pm 0.8$ & $58.4 \pm 0.8$ & $71.9 \pm 1.5$ & $56.5 \pm 0.3$ \\
    & \checkmark & \checkmark            & $46.9 \pm 1.5$ & $\mathbf{57.5} \pm 0.6$ & $20.3 \pm 1.8$ & $58.5 \pm 0.8$ & $73.4 \pm 1.2$ & $56.4 \pm 0.1$ \\
    \checkmark & \checkmark & \checkmark & $45.5 \pm 0.3$ & $\mathbf{56.7} \pm 0.7$ & $17.4 \pm 0.1$ & $57.1 \pm 1.9$ & $73.5 \pm 0.5$ & $56.6 \pm 0.2$ \\
    \bottomrule
    \end{tabular}
    }
    \caption{Performance comparison on CPA when training with learnt ToM features. F1 scores for the baseline are reported from the original paper~\cite{bara2023towards} rather than from our execution of the official code.}
    \label{app:tab:cpa_tasks}
\end{table*}

\subsubsection{GNN Plan Encoder}
We propose a modification to the method by \citet{bara2023towards} that includes representing plans as graph objects and using a GNN-based encoder-decoder together with candidate sampling to predict missing knowledge, i.e., missing edges (see \autoref{fig:model}).

In the encoding phase, given a graph, we compute the node embeddings using GATv2 convolutions~\cite{brody2022how}.
The node features (one-hot encoding of materials) are first projected using a single linear layer with hidden dimension $128$, followed by GELU activation and dropout. 
The same procedure is applied to edge features (one-hot encoding of tools). 
Then the first GATv2 convolution is applied, which has hidden dimension $128$, four heads, GELU activation and dropout. 
The final node embeddings are generated by a second GATv2 convolution with output dimension $128$ and one head. 

In the decoding phase, we evaluate potential missing edges by scoring them against the relevant node embeddings and the context vector. 
This involves combining the two node embeddings associated with the edge and the context vector, then passing the resulting concatenation through a linear layer with an output dimension of one. 
The output logits are subsequently fed into a sigmoid function with a threshold of $0.5$ to determine whether the edge exists or not.

\subsubsection{Transformer}
We use a single-block Transformer~\cite{vaswani2017attention} with output dimension $1024$ as sequence-to-sequence model. 
The input consists of concatenated features from various modalities, such as video, dialogue, plan graph, and dialogue moves (and ToM features in the case of CPA), as depicted in \autoref{fig:model}.
Our Transformer incorporates positional encoding~\cite{vaswani2017attention} and utilises a causal attention mask to ensure that each token attends only to previous tokens during self-attention computation. We utilise eight attention heads to compute attention scores over the concatenated input features, including ToM features for CPA tasks. 
This ensures that the model attends to ToM features. 
Following the baseline models~\citep{bara-etal-2021-mindcraft, bara2023towards}, our models utilise zero-padding when an input modality is absent.

\subsubsection{Training}
Models are trained using PyTorch~\citep{paszke2019pytorch} with \texttt{1}, \texttt{42} and \texttt{123} as random seeds. 
All models were trained on a single GPU card, taking approximately 60 minutes for the baselines (17,946,302 parameters) and our models (9,222,100 parameters) on ToM.
For CPA, training takes approximately 60 minutes for the baselines (33,698,691 parameters) and 20 minutes for our models (13,364,641 parameters). 
For the baselines~\citep{bara2023towards}, we used default parameters reported in the code. 
For our models, we used the Adam optimiser~\cite{kingma2015adam} with $\beta_1 = 0.9$, $\beta_2 = 0.99$, $\epsilon = 10^{-8}$, and a learning rate of $\eta = 1\cdot10^{-4}$. 
We did not perform any exhaustive hyper-parameter tuning but just tried a set of reasonable values: $\{1\cdot10^{-5}, 5\cdot10^{-4}, 1\cdot10^{-4}, 5\cdot10^{-4}\}$. 

\subsection{Comparison to \citet{bara2023towards}}
\label{app:bara}
Although we trained the baseline of \cite{bara2023towards} using the official code\footnote{\url{https://github.com/sled-group/collab-plan-acquisition}} with default hyperparameters, its performance slightly deviated from the original paper. 
We contacted the authors asking for clarifications and details that are not documented in the paper/code. 
Discussions with them did not yield a clear answer, and they were unable to provide their model files.
Therefore, in \autoref{tab:tom_tasks} and \autoref{tab:cpa_tom_feats} we provided results obtained from our runs, reproducible by using our code provided as supplementary material.
In the interest of completeness, we include a comparison between our results and those reported by \citet{bara2023towards} in \autoref{app:tab:tom_tasks} for ToM tasks and in \autoref{app:tab:cpa_tasks} for CPA tasks. 
On the ToM tasks, the scores achieved by our model are generally on par with those reported by \citet{bara-etal-2021-mindcraft}.
On CPA, our model still outperforms the baseline of \citet{bara2023towards} that, based on the originally reported scores, on average performs slightly worse in some tasks (see \autoref{app:tab:cpa_tasks}, Overall). 

It is important to node that our main contributions remain unaffected by this mismatch: 
First, our improvement on OMK stands, both if we compare our baseline results and those reported in \citet{bara2023towards}. Notably, our results for the baselines ($\sim0.28$) are higher than the ones in the original paper ($\sim0.21$).
Second, our analyses are not contingent on absolute results but rather on the relationship with the ToM tasks.

\subsubsection{Qualitative Example}
\begin{figure}[t]
    \centering
    \includegraphics[width=\linewidth]{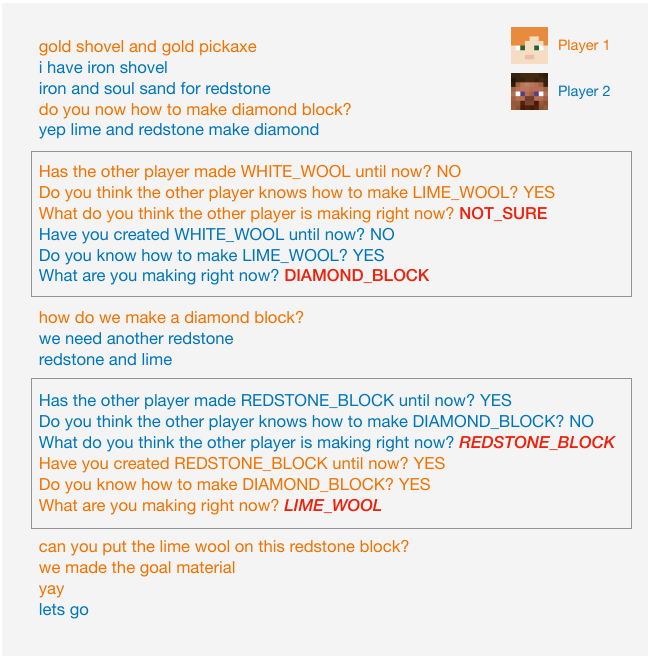}
    \caption{Chat from a game in which our ToM model cannot solve the \textit{Player Intention} ToM task. In the same game, integrating the corresponding ToM features into the CPA model enhances the performance on PMK.}
    \label{fig:chat}
\end{figure}

\autoref{fig:chat} shows the chat from a game in which our ToM model cannot solve the \textit{Player Intention} ToM task. 
However, \textit{on the same game}, integrating the corresponding ToM features into the CPA model enhances the performance on the PMK task by approximately $10$ points. 
This particular game instance is highlighted by the dotted red circle in the rightmost plot of \autoref{fig:correlation}.
We speculate the ToM model's struggle with the \textit{Player Intention} task may arise from the initial game part where players' beliefs are \textcolor{red}{misaligned}, which could result in a \textcolor{red}{\textit{false belief}} (cf. \autoref{fig:chat}).
Despite this, the CPA model still benefits from the inclusion of ToM features, suggesting that ToM models may actually be learning information that is more closely associated with other correlations in the data, rather than representing the mental states. 

\subsection{Tools}
We performed our data analysis using NumPy~\citep{harris2020array}, Pandas~\citep{reback2020pandas, mckinney2010data}, and SciPy~\citep{virtanen2020scipy}. 
Figures were made using Matplotlib~\cite{Hunter2007matplotlib}.

\subsection{Infrastructure}
We ran our experiments on a server running Ubuntu 22.04, equipped with NVIDIA Tesla V100-SXM2 GPUs with 32GB of memory and Intel Xeon Platinum 8260 CPUs.